\pdfoutput=1

\documentclass[11pt]{article}

\usepackage[final]{acl}

\usepackage{times}
\usepackage{latexsym}
\usepackage{amsmath}
\usepackage{amssymb}
\usepackage{hyperref}

\usepackage[T1]{fontenc}

\usepackage[utf8]{inputenc}

\usepackage{microtype}

\usepackage{inconsolata}
\usepackage{enumitem}
\usepackage{enumerate}
\setlist{noitemsep, topsep=0pt}
\usepackage{graphicx}
\usepackage{enumitem}

\title{Towards Open-Ended Discovery for Low-Resource NLP}

\author{\normalsize Bonaventure F. P. Dossou$^{1,2,*}$, Henri A\"idasso$^{3,*}$ \\\\
\footnotesize
$^1$McGill University $^2$Mila Quebec AI Institute, $^3$École de technologie supérieure (ÉTS)\\
\footnotesize
$^*$Equal Contribution\\
\footnotesize
bonaventure.dossou@mila.quebec, henri.aidasso@etsmtl.ca}

\begin{document}
\maketitle

\begin{abstract}
Natural Language Processing (NLP) for low-resource languages remains fundamentally constrained by the lack of textual corpora, standardized orthographies, and scalable annotation pipelines. While recent advances in large language models have improved cross-lingual transfer, they remain inaccessible to underrepresented communities due to their reliance on massive, pre-collected data and centralized infrastructure. In this position paper, we argue for a paradigm shift toward open-ended, interactive language discovery, where AI systems learn new languages dynamically through dialogue rather than static datasets. We contend that the future of language technology, particularly for low-resource and under-documented languages, must move beyond static data collection pipelines toward interactive, uncertainty-driven discovery, where learning emerges dynamically from human-machine collaboration instead of being limited to pre-existing datasets. We propose a framework grounded in joint human-machine uncertainty, combining epistemic uncertainty from the model with hesitation cues and confidence signals from human speakers to guide interaction, query selection, and memory retention. This paper is a call to action: we advocate a rethinking of how AI engages with human knowledge in under-documented languages, moving from extractive data collection toward participatory, co-adaptive learning processes that respect and empower communities while discovering and preserving the world's linguistic diversity. This vision aligns with principles of human-centered AI,
emphasizing interactive, cooperative model building between AI systems and speakers.
\end{abstract}

\section{Introduction}
The recent progress in Natural Language Processing (NLP) has been largely shaped by a data-driven paradigm. Foundation models, built on large-scale internet corpora and empowered by scaling laws, have unlocked impressive generalization across tasks and languages \cite{kaplan2020scaling, brown2020language, scao2022bloom}. However, this trajectory has come at a cost: the assumption that performance improves with ever more data and compute has made cutting-edge research increasingly inaccessible, especially to researchers and communities in the Global South \cite{sambasivan2021re, schwartz2022green}.

Despite efforts to democratize NLP, a stark imbalance persists. African languages, which make up over 30\% of the world’s linguistic diversity, account for less than 1\% of NLP research output \cite{joshi2020state}. These languages typically lack large-scale text corpora, parallel datasets, and standardized annotation practices. Transfer learning, active learning, self-supervised and semi-supervised learning, have all been proposed to address this data scarcity \cite{howard2018ulmfit, devlin2019bert, eindor2020active, dossou2022afrolm, dossou2025asr, dossou2025cf}, but even these methods depend on the availability of some unlabeled or previously seen language data. In environments where data is extremely scarce or non-digitized, such assumptions break down.

Moreover, while recent Large Language Models (LLMs) have demonstrated impressive cross-lingual abilities, their success is closely tied to data scale, computational resources, and increasingly centralized infrastructure. As scaling laws plateau and operational costs rise, the current paradigm risks becoming both \textit{unsustainable} and \textit{exclusive}, limiting participation from underrepresented communities and preventing scalable solutions for the languages that need them most \cite{strubell2019energy, bender2021dangers, pouget2024filter}.

We argue that NLP must now evolve beyond static, data-hungry training regimes. Inspired by recent work in open-ended discovery and self-improving AI \cite{hughes2024openended, siddiqui2024acd}, we propose a shift toward \textit{interactive, uncertainty-driven language learning}. In our vision, AI systems learn languages not from vast corpora, but through \textit{natural dialogue}, identifying gaps in their understanding, asking questions, and incorporating feedback in real time.

\textit{Imagine an AI system that only understands English, but receives a human input in Fon \cite{dossou-emezue-2020-ffr, dossou-emezue-2021-okwugbe, dossou2021afrivec, dossou2021crowdsourced, dossou2023fonmtl}. Instead of guessing or ignoring it, the system responds: \textbf{\textit{``I do not recognize this language. Could you help me understand it?''}} From this first exchange, it starts acquiring the new linguistic concepts interactively. Over time, through repeated exposure and correction, the system transitions from total ignorance to conversational fluency in the new language. This vision shifts the emphasis \textbf{from training on what we have to} learning from \textbf{what we do not yet understand}, as humans do.}

In this position paper, we explore the technical and conceptual foundations for such systems. We argue that \textit{open-ended language learning}, grounded in \textit{epistemic uncertainty}, \textit{dialogue}, and \textit{human-in-the-loop adaptation}, represents a scalable and inclusive path forward for low-resource NLP, especially in contexts where static data is not available, representative, or sufficient. We also outline a set of \textit{open challenges} that arise from this vision, including the need for reliable uncertainty estimation, continuous learning mechanisms, and equitable access to interaction data. We discuss both the promise and the risks of this approach, including the question of whether such systems can acquire meaningful language competence without sufficient exposure or human feedback, and what architectures, incentives, or evaluation schemes would be required to support them.

\section{Background and Related Work}

\subsection{Low-Resource Languages}
\label{subsec:low_resource_languages}

Africa is one of the most linguistically diverse continents, home to over 3,000 indigenous languages \cite{epstein_language_1998, ethnologue27}, which account for about one-third of the world's 7,159 living languages \cite{ethnologue27}. In an increasingly digital world, where today’s AI advancement such as LLMs offer unprecedented possibilities, the non-integration of these languages into the technological landscape not only exacerbates social inequalities but also poses a serious threat to the survival of entire linguistic cultures. 

As inclusion and diversity gain global importance, commendable efforts have been made by researchers to identify available, albeit scarce, data sources (e.g., the Bible in Fon). Moreover, there are growing efforts for datasets creation (sometimes done manually and on a voluntary basis). These datasets have been used to create machine translation models that produce acceptable results \cite{dossou-emezue-2020-ffr,adelani-etal-2022-thousand}. As a result, some of the very low-resource languages such as Fon, Ewe have been recently integrated into Google Translate,\footnote{\url{https://translate.google.com/?sl=en&tl=fon}} for textual translations.

Despite these important advances, several major challenges persist that existing solutions do not, and arguably cannot address. In particular, current approaches still rely heavily on larger amounts of textual data \cite{adelani-etal-2022-thousand,dossou2022afrolm,nekoto-etal-2020-participatory}, resources that are extremely scarce or absent for many African languages and dialects \cite{nekoto-etal-2020-participatory,joshi2020state}. Due to this reliance, existing solutions only cover a tiny fraction ($\approx$1\%) of the languages, typically selected based on speaker population size or researchers' ties \cite{adelani-etal-2022-thousand,adelani-etal-2022-masakhaner}. These choices overlook the existing diversity and will ineluctably reinforce existing social inequalities and discrimination. For instance, Nigeria alone has over 500 indigenous languages \cite{ethnologue27}, most of which severely lack written resources. Even more concerning is the practical impact of current solutions. In fact, most low-resource languages exist solely through oral traditions, meaning that the vast majority of native speakers can only speak them and struggle to read written versions, if such versions exist at all \cite{dossou-emezue-2021-okwugbe,afrispeech,afrinames}. Therefore, solutions that rely on textual translations are fundamentally misaligned with how these languages are actually used, making them ineffective for real-world communication needs.

\subsection{Human Uncertainty Estimation}
\label{subsec:human_uncertainty}

Incorporating human uncertainty into interactive learning frameworks has emerged as a critical complement to model uncertainty, as human feedback is often non-deterministic and can significantly shape model learning dynamics. \citet{collins2023conceptuncertainty} explore concept-level interventions where humans provide feedback on intermediate concepts rather than final labels. They show that capturing the \textit{confidence} or uncertainty of these interventions, through soft labels or probabilistic feedback, improves model robustness and generalization.

\citet{mendes2025humanmodeluncertainty} study the relationship between human-perceived and model-predicted uncertainties, finding only limited correlation between the two. This indicates that model uncertainty alone is insufficient to assess ambiguity in real-world settings. Explicitly modeling human uncertainty, for example, through elicited confidence scores or inter-annotator variance, can lead to more calibrated and reliable learning.

From a broader perspective, \citet{bhatt2020uncertaintytransparency} argue that exposing both human and model uncertainties enhances transparency and mutual understanding in human-AI collaboration. Similarly, collaborative annotation frameworks such as CoAnnotating~\cite{coannotating2023} leverage these uncertainty estimates to decide when to defer to human expertise or proceed autonomously, improving both efficiency and reliability in human-in-the-loop learning pipelines.

\begin{figure*}[htbp]
\centering
      \includegraphics[width=\linewidth]{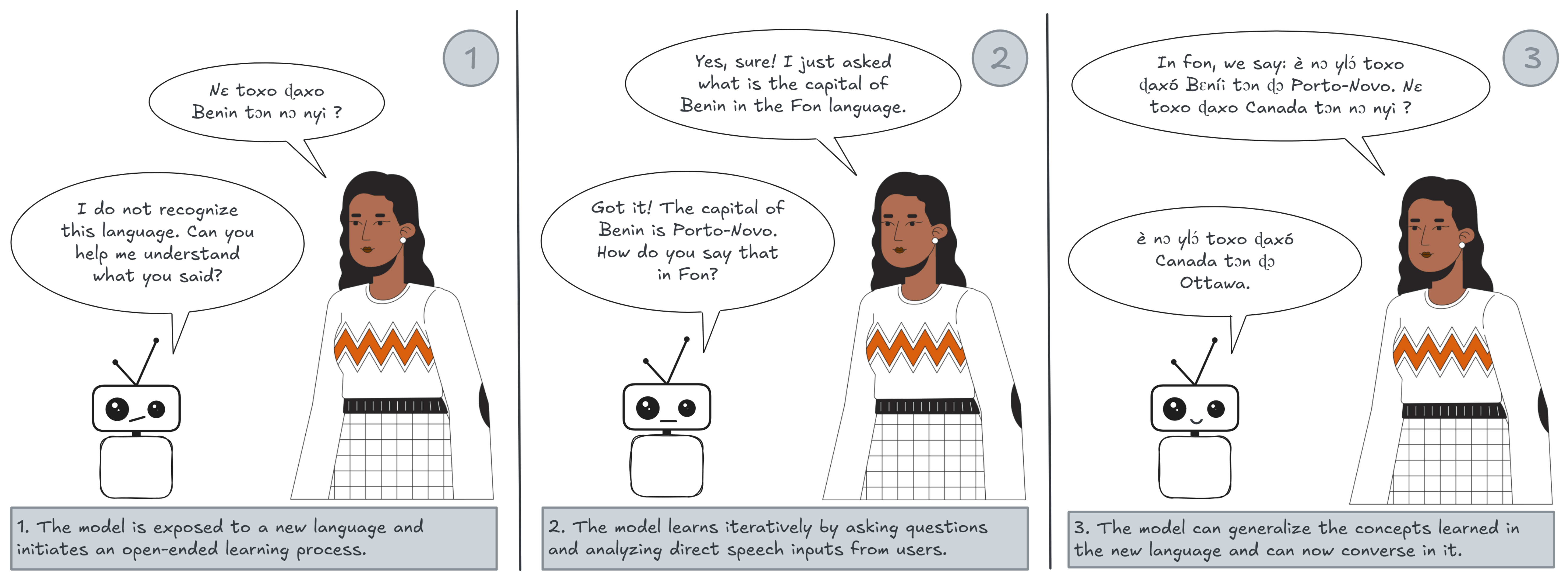}
\caption{Illustration of the proposed approach for open-ended learning of low-resource languages. It shows the voice conversation between a human and an agent who teaches the agent to recognize and respond to requests for the capital city of a country in the Fon language.}
\label{fig:open_ended_learning}
\end{figure*}

\subsection{Model Uncertainty Estimation}

In machine learning models, uncertainty estimation plays a crucial role in determining whether a model can respond confidently or should request clarification from the user. We denote by \( f_\theta: \mathcal{X} \rightarrow \mathcal{Y} \) a parametric model with parameters \(\theta\), input \(x \in \mathcal{X}\), and predictive distribution \(p_\theta(y|x)\). $\mathcal{D}$ is the training dataset.

\citet{kendall2017uncertainties} distinguish two types of uncertainty: aleatoric uncertainty (\(U_a\)) and epistemic uncertainty (\(U_e\)). Most literature works focus on \(U_e\) which is approximated by:
    \[
    U_e(x) = \mathbb{V}_{p(\theta|\mathcal{D})}[\mathbb{E}_{p(y|x,\theta)}[y]]
    \]

This is \(U_e\) because directly tied to limited data or lack of model knowledge.
The two most common ways of estimating \(U_e\) are the following:
\paragraph{With Bayesian Neural Networks}
BNNs \cite{mackay1992practical,neal1996bayesian} define a posterior over weights:
\[
p(\theta|\mathcal{D}) \propto p(\mathcal{D}|\theta) p(\theta),
\]
and predictive uncertainty as:
\[
p(y|x,\mathcal{D}) = \int p(y|x,\theta)p(\theta|\mathcal{D})\,d\theta.
\]
In practice, this integral is untractable, and approximated using variational inference \cite{blundell2015weight} or Monte Carlo sampling \cite{gal2016dropout}.

\paragraph{With Deep Ensembles}
Given \(M\) independently trained models \(\{f_{\theta_m}\}_{m=1}^M\), predictive uncertainty is quantified via:
\[
U_m(x) = \frac{1}{M}\sum_m H(f_{\theta_m}(y|x)),
\]
where \(H(\cdot)\) is Shannon entropy.

In summary, while advances in uncertainty estimation have improved model reliability~\cite{kendall2017uncertainties, gal2016dropout, kirsch2019batchbald} and recent work has explored uncertainty in human feedback~\cite{collins2023conceptuncertainty, mendes2025humanmodeluncertainty}, current AI systems still learn predominantly from static datasets or treat user input as deterministic corrections. This creates two major limitations: (i) uncertainty from humans and models is rarely considered jointly, reducing the system's ability to assess when to seek clarification or defer decisions, and (ii) learning processes remain largely offline, without mechanisms to dynamically adapt to evolving user input.

To address these shortcomings, we introduce an \textbf{interactive learning system} that moves beyond passive, data-driven training. Instead of relying solely on pre-collected corpora, the system engages directly with users, identifies gaps in understanding, requests clarification when uncertainty is high, and incorporates feedback into its evolving knowledge state. This approach aims to fuse human and model uncertainties to guide the dialogue flow, enabling real-time, adaptive, and more sample-efficient language acquisition.

\section{Proposed Approach}
Our proposed framework enables AI systems to acquire language competence through open-ended, interactive learning. This process is illustrated in Figure \ref{fig:open_ended_learning} through interactions between a human and the AI system (agent).
Rather than training on large static corpora, the system learns by engaging with users in real time, identifying gaps in its knowledge, soliciting clarification, and integrating feedback. The methodology consists of three core components: (1) modeling interactional uncertainty, (2) language acquisition via feedback, and (3) continual learning from dialogic exposure.

\subsection{Modeling Interactional Uncertainty}
\label{stage1}
At the heart of our approach is the notion of epistemic uncertainty, which refers to the system's awareness of what it does not know. In conventional NLP, model uncertainty is often used for tasks like active learning or confidence calibration~\cite{kendall2017uncertainties, gal2016dropout, houlsby2011bayesian, guo2017calibration}. Here, we extend this principle to guide decision-making during interactive language learning.

We define a composite uncertainty signal combining both human and machine contributions:
\[\mathcal{U}_{\text{total}} = \alpha \cdot \mathcal{U}_{\text{human}} + (1 - \alpha) \cdot \mathcal{U}_{\text{model}}\]
where $\mathcal{U}_{\text{model}}$ is the model's epistemic uncertainty, estimated via entropy, ensemble disagreement, or Bayesian approximations~\cite{kendall2017uncertainties,kirsch2019batchbald,bald,gal2016dropout,kirsch2023stochasticbatchacquisitionsimple}, $\mathcal{U}_{\text{human}}$ reflects uncertainty inferred from hesitation cues, conflicting corrections, or prosodic markers, and $\alpha$ controls the relative influence of human versus machine uncertainty.

Given this signal, the system selects a query $Q^*$ to ask the human speaker, optimizing:
\[
Q^* = \arg\max_Q \frac{\mathbb{E}[\text{InfoGain}(Q)]}{\text{Cost}(Q, \mathcal{U}_{\text{human}})}
\]
where 
\[
\text{Cost}(Q, \mathcal{U}_{\text{human}}) = c(Q)\big(1+\lambda *\mathcal{U}_{\text{human}}\big)
\]
with $c(Q)$ representing the baseline time or cognitive effort required for query type $Q$, and $\lambda \geq 0$ controlling how strongly human uncertainty increases perceived cost. This interaction cost reflects the human effort required to answer a query and the likelihood of confusion when the speaker is already uncertain. Scaling the cost by $(1+\lambda *\mathcal{U}_{\text{human}})$ ensures the system avoids queries that are both expensive and likely to yield ambiguous responses. This improves efficiency and user experience, making learning cooperative rather than extractive.

The expected information gain from a query $Q$ is defined as the anticipated reduction in predictive uncertainty:
\[
\begin{split}
\text{InfoGain}(Q) &= \mathbb{H}[Y \mid x, \mathcal{D}] \\
&\quad - \mathbb{E}_{A \sim p(A \mid Q)}
   \big[\mathbb{H}[Y \mid x, \mathcal{D}, Q, A]\big]
\end{split}
\]
where $\mathbb{H}[\cdot]$ denotes Shannon entropy, $\mathcal{D}$ is the current learner state, and $A$ denotes a human response sampled from $p(A \mid Q)$. This term quantifies how much uncertainty the query is expected to resolve. We define $p(A|Q)$ as the conditional distribution over possible human responses given a query $Q$. This distribution models the variability and uncertainty in human feedback due to ambiguity in meaning, hesitation or noise in responses, and contextual variability across speakers.

The selected query $Q^*$ and the anticipated distribution of human responses $p(A|Q)$ provide the necessary context for the next stage, where human feedback is integrated into the model.

\subsection{Language Acquisition via Human Feedback}
\label{stage2}
Once a query $Q^*$ has been selected based on the joint uncertainty signal, the AI system receives a feedback signal $A$ from the human speaker. In this stage, the goal is to integrate the new information into the model’s knowledge while accounting for both human and model uncertainty.

A targeted query $Q$ is designed to elicit clarifying information about input $x$, such as asking \textbf{\textit{``What does this word mean?''}} or \textbf{\textit{``How would you say this sentence?''}}. The response is denoted as $A \sim p(A|Q)$, sampled from a conditional distribution over possible answers. This distribution reflects that feedback may vary or include ambiguity, such as multiple possible translations or uncertain corrections.

We denote $p_\theta(\cdot|x)$ as the model's current predictive distribution over possible meanings or utterances for input $x$, parameterized by $\theta$. The human feedback is represented as $y_{\text{human}}$, a meaning distribution derived from the response $A$. It can be sharp, corresponding to a single unambiguous answer, or soft, capturing several plausible meanings with associated probabilities. Finally, we introduce a reliability weight $w_f = 1-\mathcal{U}_{\text{human}}$, which downscales the influence of uncertain human feedback. When human uncertainty is high, the system places less emphasis on the feedback to avoid reinforcing potentially misleading signals.

Using these definitions, the system constructs a new target distribution that combines its own prior predictions with the received feedback:
\[
\tilde{y} = w_f \cdot y_{\text{human}} + (1-w_f) \cdot p_\theta(\cdot|x)
\]
This weighted target guides the parameter update:
\[
\theta' = \theta - \eta \nabla_\theta \mathcal{L}(p_\theta(\cdot|x), \tilde{y})
\]
where $\eta$ is the learning rate and $\mathcal{L}$ is a loss function. KL Divergence can be used to align the model's predicted distribution with human-provided meaning probabilities in a continuous space, making it well-suited for uncertain or soft feedback. Contrastive Loss distinguishes correct meanings from alternative ones in an embedding space, supporting open-ended discovery where meanings are not predefined. Categorical Cross-Entropy works when the system has a finite set of candidate meanings, though it is less ideal for open-ended language learning since it assumes predefined categories.

This approach allows the system to integrate human feedback incrementally and proportionally to its reliability, while still preserving useful prior knowledge from its own predictions. In the future, more appropriate loss functions could be designed specifically for dialogic, open-ended learning scenarios to better reflect the uncertainty and flexibility inherent in human language interactions.

\subsection{Continual Learning from Dialogic Exposure}
\label{stage3}
Language acquisition is not a single-step process. Over multiple interactions, the system must consolidate knowledge, refine uncertain examples, and adapt to evolving feedback. To achieve this, every interaction is stored in a memory bank:
\[
\mathcal{M} = \{(x_i, A_i, w_i)\}
\]
where each element consists of the input $x_i$, the human feedback $A_i$, and an associated weight:
\[
w_i = (1-\mathcal{U}_{\text{human}}^{(i)})(1-\mathcal{U}_{\text{model}}^{(i)})
\]
This weight captures the combined confidence of both the human and the model for a given interaction.

The memory bank $\mathcal{M}$ acts as a growing repository of past interactions with human speakers, each stored alongside a weight indicating reliability. Periodically, the system revisits stored samples to reinforce reliable information and re-query ambiguous examples. Past interactions are used to improve the model through uncertainty-aware gradient updates:
\[
\theta \leftarrow \theta - \eta \sum_i w_i \nabla_\theta \mathcal{L}(p_\theta(\cdot|x_i), A_i)
\]

\noindent $p_\theta(\cdot|x_i)$ refers to the same predictive model introduced in Section~\ref{stage2}, now updated iteratively using both immediate feedback and stored memory samples. We reuse the notation to emphasize that the model evolves over time through repeated uncertainty-guided interactions.

Low-weight samples contribute less to the update, preventing uncertain or noisy feedback from degrading the learned representation. They are not discarded but flagged for future re-querying when opportunities arise. This creates a closed interactive loop where the system encounters new input $x$, computes $\mathcal{U}_{\text{total}}$ and selects an optimal query $Q^*$, collects human feedback $A$ and updates parameters incrementally, stores the interaction in $\mathcal{M}$ with weight $w_i$, and periodically revisits uncertain cases to refine or validate earlier knowledge, looping back when necessary.

Through these mechanisms, uncertainty evolves from a static confidence score into an active principle governing when to trust, query, defer, or memorize. This continual process ensures that learning is incremental, reliable, and co-adaptive. It enables the system to refine its internal representations over time, progressively improving its understanding of a new language while remaining sensitive to the reliability of past and future feedback. Together, these three stages establish a self-reinforcing loop for interactive language discovery, where uncertainty not only shapes individual interactions but also drives long-term, co-adaptive learning.

\section{Opportunities and Challenges}
Our proposed framework for open-ended language discovery leverages joint human-machine uncertainty to guide interaction, query selection, and memory retention. While the approach introduces a novel paradigm for low-resource language acquisition, its success and limitations stem directly from the mechanisms we designed. Unlike conventional NLP pipelines that rely on static, curated datasets and post-hoc analysis, this framework is designed for real-time, adaptive interaction. It emphasizes uncertainty-driven decision-making, enabling language acquisition to progress even when large corpora, standardized orthographies, or expert annotators are unavailable.

\subsection{Why This Could Work}

The framework builds on several principles that make it uniquely suited for interactive, low-resource settings. By explicitly modeling epistemic uncertainty, the system learns what it does not know and can focus queries on areas of high information gain rather than engaging in blind memorization. This targeted querying mechanism has the potential to accelerate language acquisition compared to static corpus-based training approaches. Incorporating $\mathcal{U}_{\text{human}}$ allows the system to defer or prioritize information based on human confidence, ensuring that reliable feedback from fluent speakers directly shapes the learned representation and reduces noise in the earliest stages of learning. Over time, dynamic weighting ($\alpha$) adapts reliance on each contributor according to their observed consistency and reliability, making the system robust to heterogeneous or occasional feedback. Furthermore, confidence-weighted memory retention enables iterative refinement of knowledge: high-certainty information consolidates quickly, while ambiguous examples remain open for re-querying, progressively building a stable and trustworthy knowledge base. Together, these mechanisms enable data-efficient learning that can bootstrap language understanding from a small number of high-value interactions, making it feasible in settings where large corpora are unavailable. These properties suggest that joint human–machine uncertainty could form the backbone of scalable, respectful, and data-efficient language acquisition, where conventional supervised NLP pipelines cannot operate.

\subsubsection{In the Context of Low-Resource African Languages}

Low-resource African languages often face a unique combination of challenges that make standard NLP pipelines ineffective: severe data scarcity, highly variable orthographies, oral traditions without standardized writing systems, and limited availability of expert annotators. The proposed framework is particularly well-suited to this context because it does not rely on pre-existing corpora or formal linguistic resources. Instead, it learns interactively from small, high-value exchanges, asking only those questions that are most informative given its current uncertainty. This targeted learning process minimizes the burden on speakers, who may have limited time or literacy in standardized orthography, while still allowing the system to rapidly form hypotheses about grammar, semantics, and phonology.

Moreover, the joint modeling of human and machine uncertainty makes the framework robust to the realities of field data collection in African settings, where contributors may have varying degrees of fluency, confidence, or even differing dialects of the same language. By adapting reliance on each contributor through dynamically learned weighting ($\alpha$), the framework can filter noise while still capturing dialectal richness. Its ability to defer uncertain information and revisit ambiguous examples ensures that rare or culturally significant linguistic forms are not prematurely discarded. These properties make it a promising approach for preserving, documenting, and learning African languages where the cost of traditional data collection is prohibitive and where respectful, participatory collaboration with speakers is essential. This approach not only addresses data scarcity but also reframes language technology development as a collaborative process between AI systems and speakers. By moving away from extractive data collection toward live, adaptive interaction, it offers a pathway for NLP to support language documentation and revitalization efforts. Particularly in marginalized communities, this paradigm empowers speakers to co-create technology aligned with their linguistic and cultural realities, potentially reshaping how AI contributes to the preservation and expansion of global linguistic diversity.

\subsubsection{In the Context of Human-Centered AI and Human-Computer Interactions}

The proposed framework embodies principles of human-centered artificial intelligence by placing speakers at the center of the learning process. Rather than treating them as static annotators or sources of labels, it engages in a cooperative interaction where both human and machine uncertainty guide the flow of information exchange. This fosters transparency and trust, as speakers can see that the system acknowledges its own uncertainty, adapts to their confidence levels, and defers decisions when information is unclear.

From a Human-Computer Interaction (HCI) standpoint, the framework reduces the cognitive and emotional burden on contributors by focusing only on high-value, contextually relevant questions instead of overwhelming them with repetitive or trivial requests. It can adapt the pace and style of interaction based on hesitation cues, feedback latency, or non-verbal indicators of uncertainty, making it more accessible to non-expert participants. Additionally, the iterative refinement of memory ensures that early mistakes can be revisited and corrected collaboratively, giving speakers a sense of agency and ownership in shaping the emerging language model. This paradigm transforms data collection from a one-way, extractive process into a participatory dialogue, contributing to the development of AI systems that are not only technically effective but also socially aligned and respectful toward the communities they aim to serve. In doing so, it demonstrates a path toward genuinely human-centered AI, where computational methods adapt to people, rather than asking people to adapt to technology. This vision is aligned with participatory and co-design approaches explored in HCI research \cite{liao2023ai,birhane2022power,10.1145/3617694.3623261}, which emphasize collaborative model building, transparency, and community agency in shaping AI behavior.

While these properties highlight the potential of our framework to enable scalable, and data-efficient language learning, realizing this vision in practice is far from trivial. Uncertainty-guided discovery introduces its own vulnerabilities, and deploying such systems in real-world low-resource environments presents additional technical and sociotechnical barriers, that must be addressed. The following section discusses these open challenges.

\subsection{Challenges}
Several challenges could undermine the effectiveness of the proposed framework in practice. A first concern lies in the reliability of uncertainty estimation. Because the system operates on highly out-of-distribution data such as new languages, unseen constructs, and unpredictable input patterns, its uncertainty signals may not be well calibrated. Miscalibration could lead to redundant or unnecessary queries, or conversely, to missed opportunities to acquire valuable information early on.

Human uncertainty signals introduce another layer of complexity. Hesitation cues, conflicting answers, or silence are not always reliable indicators of a speaker’s true confidence. Cultural norms and individual communication styles can further distort these signals, leading the system to over-trust uncertain information or defer excessively even when a speaker would have provided correct input. This unreliability in feedback interpretation can propagate downstream errors in learning.

Errors may also arise in the adaptive weighting mechanism. Because $\alpha$ must be learned online from sparse observations, early interactions can dominate future weighting, allowing biases from the first few contributors to persist unchecked. In heterogeneous communities where speaker reliability varies widely, it becomes difficult to estimate contributor trustworthiness accurately, which risks amplifying noise and reducing the value of human input. This interacts closely with query selection: without stable reliability or cost estimates, the system may waste interactions on poorly chosen clarifications, frustrating users and slowing overall progress.

The memory component presents its own risks. Confidence-weighted retention is designed to consolidate reliable information quickly, but if misinterpreted feedback is assigned high confidence, early errors risk becoming fossilized in the learned representation. Conversely, rare linguistic forms may repeatedly receive low-confidence scores, preventing their integration and leaving parts of the language undocumented or misunderstood. This challenge is compounded in what we term a “double-uncertainty deadlock,” where both the model and the human contributors remain uncertain for extended periods. In such cases, the system may repeatedly defer decisions, becoming overly cautious and failing to test hypotheses that could break the cycle of uncertainty.

Finally, practical constraints in real-world deployment cannot be ignored. Reliable uncertainty estimation, adaptive weighting, and dynamic query selection all introduce computational overhead that may be infeasible on low-cost, battery-limited, or offline devices. Connectivity issues, limited processing power, and fragile hardware environments could hinder the ability of the framework to operate effectively in the very settings it aims to serve.

\subsection{Future Directions}
Addressing these challenges requires progress on several fronts. Improving epistemic uncertainty estimation in open-ended, out-of-distribution language input is a priority, as more reliable measures would reduce unnecessary queries and strengthen the system's ability to make informative decisions early on. Equally important is the development of context-aware and culturally adaptive models of human uncertainty, since hesitation and confidence cues vary widely across individuals and communities. Advancing methods for learning $\alpha$ from sparse interactions will also be key to mitigating early biases, ensuring that the system adapts fairly and dynamically to multiple contributors over time.

Meta-learning approaches offer a promising path toward improving $\alpha$ estimation. By transferring priors on speaker reliability from related language acquisition sessions or typologically similar languages, the system could begin with more informed weighting strategies, reducing the risk of overfitting to a handful of early interactions. This would make adaptation faster and more stable, even in diverse or previously unseen linguistic settings.

Developing multi-agent exploration policies could further enhance query selection. Instead of treating human interactions in isolation, coordinated strategies could balance information gain, contributor reliability, and annotation cost across multiple speakers. Such strategies might deliberately diversify queries to capture rare linguistic forms, seek cross-validation from independent sources to resolve ambiguities, and avoid overloading single contributors, making learning more efficient and collaborative.

Breaking double-uncertainty deadlocks will require exploration mechanisms that take calculated risks when both human and model uncertainty remain high. Periodic re-querying, rediscovery routines, and targeted hypothesis testing could help overcome conservativeness and expand the system's knowledge base over time. Finally, lightweight, offline-capable implementations of the framework are necessary for real-world deployment. Achieving efficient uncertainty estimation, adaptive query selection, and meta-learning-based weighting on low-power devices would make the approach scalable and practical for under-resourced communities that lack access to high-compute infrastructure.

If these research directions are pursued, joint human-machine uncertainty could unlock scalable, interactive, and respectful language learning systems capable of discovering and documenting under-resourced languages without relying on large curated datasets. Ultimately, this line of research bridges technical innovation and participatory design, opening opportunities for AI systems that learn with people, not just from data.

\section{Conclusion}
This paper outlines a vision for open-ended language discovery based on joint human-machine uncertainty. We argue that future NLP systems, particularly for low-resource languages, must move beyond static data pipelines and toward interactive, participatory approaches that adapt to sparse, uncertain, and heterogeneous feedback. While many technical and sociotechnical challenges remain, this is not merely a research proposal but an ideological stance: language technology should be co-created with speakers. We position this work as a challenge to current practices that treat language as extractable data, advocating instead for AI systems that become collaborative participants in language preservation and revitalization. We call on the NLP and HCI research communities to develop methods, tools, and evaluation practices that support co-adaptive language learning systems, opening new pathways for linguistic documentation, preservation, and empowerment in the digital age. This position paper advocates for a paradigm shift: from building models that passively learn from existing data to designing systems that actively learn with people in real time, fostering respectful, human-centered AI for linguistic diversity.

\bibliography{references}
\end{document}